\documentclass{article}

    \PassOptionsToPackage{numbers, compress}{natbib}

    \usepackage[preprint]{neurips_2022}

\usepackage[utf8]{inputenc} 
\usepackage[T1]{fontenc}    
\usepackage{hyperref}       
\usepackage{url}            
\usepackage{booktabs}       
\usepackage{amsfonts}       
\usepackage{nicefrac}       
\usepackage{microtype}      
\usepackage{xcolor}         

\usepackage{graphicx} 
\usepackage{wrapfig}
\usepackage{amsmath}

\usepackage{float}
\usepackage{comment}
\definecolor{MyBlue}{HTML}{C6E9FF}
\definecolor{MyGreen}{HTML}{D6FDC3}
\definecolor{AppendixViolet}{HTML}{9999FF}
\definecolor{AppendixBlue}{HTML}{99CCFF}
\definecolor{AppendixGreen}{HTML}{CCFF99}
\definecolor{AppendixYellow}{HTML}{FFFF99}
\definecolor{AppendixPink}{HTML}{FFAAFF}
\usepackage{enumerate}

\title{CogVideo: Large-scale Pretraining for Text-to-Video Generation via Transformers}

\author{%
  Wenyi Hong\textsuperscript{\textdagger}\thanks{Equal contribution.} \ \ Ming Ding\textsuperscript{\textdagger}\footnotemark[1] \ \ Wendi Zheng\textsuperscript{\textdagger} \ \ Xinghan Liu\textsuperscript{\textdagger} \ \ Jie Tang\textsuperscript{\textdagger}\textsuperscript{$\ddagger$}\\
  \textsuperscript{\textdagger}Tsinghua University\ \ \textsuperscript{$\ddagger$}BAAI\\
  \texttt{\{hongwy18@mails, dm18@mails, jietang@mail\}.tsinghua.edu.cn} \\
}

\begin{document}

\maketitle


\begin{abstract}
Large-scale pretrained transformers have created milestones in text (GPT-3) and text-to-image (DALL-E and CogView) generation. Its application to video generation is still facing many challenges:
The potential huge computation cost makes the training from scratch unaffordable;
The scarcity and weak relevance of text-video datasets hinder the model understanding complex movement semantics. In this work, we present 9B-parameter transformer CogVideo, 
trained by inheriting 
a pretrained text-to-image model, CogView2. We also propose multi-frame-rate hierarchical training strategy to better align text and video clips. As (probably) the first open-source large-scale pretrained text-to-video model,  CogVideo outperforms all 
publicly available models at a large margin in machine and human evaluations. 
\end{abstract}

\begin{figure}[H]
  \centering
  \vspace{-5mm}%
  \includegraphics[width=\textwidth]{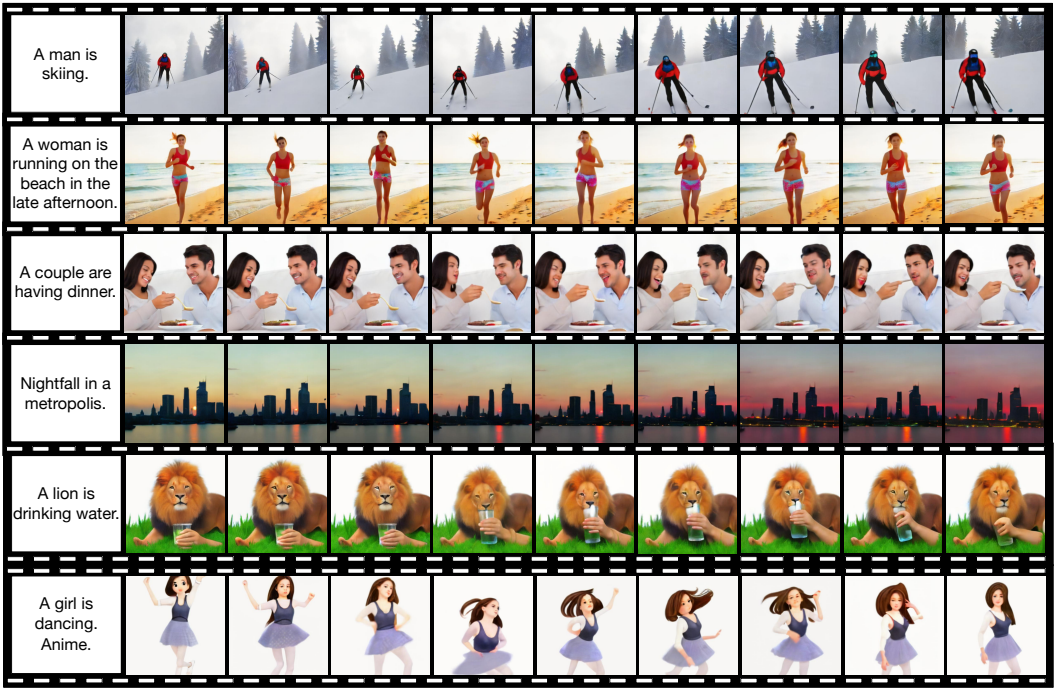}
  \vspace{-3mm}%
  \caption{Samples generated by CogVideo. The actual text inputs are in Chinese. Each sample is a 4-second clip of 32 frames, and here we sample 9 frames uniformly for display purposes. More samples, models and codes will be available at \url{https://github.com/THUDM/CogVideo}.}
  \label{fig:samples}
\end{figure}

\section{Introduction}\label{sec:intro}

Autoregressive transformers, e.g. DALL-E~\cite{ramesh2021zero} and CogView~\cite{ding2021cogview}, have revolutionized text-to-image generation recently. It is natural to investigate the potential of autoregressive transformers on text-to-video generation. Previous works followed this basic framework~\cite{wu2021n,ge2022long}, e.g. VideoGPT~\cite{yan2021videogpt}, verifying its superiority over GAN-based methods~\cite{clark2019adversarial,tulyakov2018mocogan}, but are still far from satisfactory. 

One common challenge is that the generated video frames tend to gradually deviate from the text prompt, making the generated characters hard to perform the desired actions. Vanilla autoregressive models might be good at synthesizing videos with  regular (e.g. straightly moving cars) or  random patterns (e.g. speaking by randomly moving lips), 
but fail on text prompt such as ``a lion is drinking water''.
The main difference between the two cases is that, in the former case the first frame already provides sufficient information for the subsequent changes, while in the latter  the model has to precisely understand the action ``drink'' in order to correctly generate the desired action --- the lion lifts the glass to its lip, drinks and then puts down the glass.

Why do the autoregressive transformers well understand the text-image relations, but struggle to understand the text-action  relations in videos? We hypothesize that the datasets and the way to utilize them are the main reasons.

First, it is possible to collect billions of high-quality text-image pairs from Internet~\cite{ramesh2021zero}, but the text-video data are more scarce. The largest annotated text-video dataset, VATEX~\cite{wang2019vatex},  has only 41,250 videos. The retrieval-based text-video pairs, e.g. Howto100M~\cite{miech2019howto100m}, are weakly relevant and most of them only describe the scene without the temporal information. 

Second, the duration of videos varies a lot. Previous models split the video into many clips of a fixed number of frames for training, which 
destroys the alignment between the text and its temporal counterparts in the video. If a ``drinking'' video is split into four individual clips of ``holding a glass'', ``lifting'', ``drinking'' and ``putting down'' with the same text ``drinking'', the model will be confused to learn the accurate meaning of drinking. 

\textbf{Present Work. }Here we present a large-scale pretrained text-to-video generative model, CogVideo, which is of 9.4 billion parameters  and trained on 5.4 million text-video pairs. We build  CogVideo  based on a pretrained text-to-image model, CogView2~\cite{ding2022cogview2}, in order to inherit the knowledge learned from the text-image pretraining. To ensure the alignment between text and its temporal counterparts in the video, we propose the \emph{multi-frame-rate hierarchical training}. The flexibility of the textual condition makes it possible to simply prepend a piece of text describing the frame rate to the original text prompt for modeling different frame rates. To keep the text-video alignment, we  choose a proper frame rate description to include the complete action in each training sample. The frame rate token also controls the intensity of the changes throughout continuous frames in generation. Specifically, we train a sequential generation model and a frame interpolation model. The former model generates key frames according to the text, and the latter recursively fill the middle frames by varying the frame rates to make the video coherent. As shown in Figure~\ref{fig:samples}, CogVideo can generate high-resolution (480$\times$480) videos. Human evaluation demonstrates that CogVideo outperforms all publicly available models at a large margin. 
Our main contributions can be concluded as follows:

\begin{itemize}
    \item We present CogVideo, which is 
    the \textbf{largest} and \textbf{the first open-source} pretrained transformer for text-to-video generation in the general domain. 
    \item CogVideo 
    elegantly and efficiently finetunes a pretrained text-to-image generative model for text-to-image generation, avoiding the expensive full pretraining from scratch.
    
    \item We propose the multi-frame-rate hierarchical training to better align text-clip pairs, which significantly improves the generation accuracy, in particular for movements of  complex semantics. This training strategy 
    endows CogVideo with the capacity of controlling the intensity of changes during the generation. 
\end{itemize}

\section{Related Work}

\subsection{Video Generation}
Video generation is a long-standing research topic. Most previous works 
focus on the next-frame prediction task --- forecasting the future frames based on the first video frame. Early works, e.g. CDNA~\cite{finn2016unsupervised} and PredRNN~\cite{wang2017predrnn}, leverage deterministic methods to directly predict the next frame via CNNs or RNNs.
However, these deterministic models are unable to capture the stochastic temporal patterns and synthesize coherent complex scenes. 
Generative models, especially  Generative Adversarial Networks~\cite{goodfellow2014generative} (GANs), begin to dominate the area
as they can perform unconditional or class-conditional video synthesis without the first frames. 
VGAN~\cite{vondrick2016generating} is the first one
to use GAN for video generation. It decomposes video to a static background and a moving foreground, and then generates them with 2D and 3D convolutional networks respectively. TGAN\cite{saito2017temporal} proposes to separately generate the temporal latent variables and spatial information, and MoCoGAN~\cite{tulyakov2018mocogan} similarly decomposes the latent space into context and motion subspaces. DIGAN~\cite{yu2022generating} applies implicit neural representations for video encoding. 
Recently, text-to-video generation emerges as a promising direction. The framework of VQVAE~\cite{van2017neural} and autoregressive transformers~\cite{vaswani2017attention,brown2020language} quickly becomes the mainstream method~\cite{wu2021godiva, wu2021n, ge2022long}. \citet{ho2022video} proposes video diffusion model along with a gradient method recently for text-to-video generation. The previous methods are basically trained on a specific dataset, e.g. UCF-101~\cite{soomro2012ucf101}, making the trained model domain-specific. 
Moreover, most of these models are not publicly available. 

\subsection{Autoregressive Transformer}
Recent years have witnessed the autoregressive transformer emerging as a powerful generative model. The autoregressive models 
become the most prevalent framework for text generation~\cite{sutskever2011generating}. With its prominent capacity of fitting, transformer~\cite{vaswani2017attention} gradually becomes the standard neural structure for text generation. 
One milestone is GPT-3~\cite{brown2020language}.
In computer vision,~\citet{van2017neural} 
first proposes to train a VQVAE to compress the image into a sequence of tokens from a learned dictionary, which can be efficiently handled by autoregressive models. VQ-GAN~\cite{esser2020taming} learns a more semantic-aware dictionary for unconditional image generation.  
In the text-to-image generation, pretrained autoregressive transformers such as DALL-E~\cite{ramesh2021zero} and CogView~\cite{ding2021cogview} have shown 
superiority in open-domain image generation. Besides the pure GPT-style generation, 
CogView2~\cite{ding2022cogview2} proposes a new language model CogLM for infilling in the image generation.

Recent autoregressive transformers~\cite{rakhimov2020latent, yan2021videogpt, wu2021godiva, wu2021n} have also shown their superiority in video generation. 
Among them, GODIVA~\cite{wu2021godiva} and NÜWA~\cite{wu2021n} focus on the open-domain text-to-video generation.
However, they simply generate frames or frame blocks one by one in a chronological order, and may suffer from poor text-video alignment (Cf. \S~\ref{sec:intro}). 


\section{Method}

\begin{figure}
  \centering
  \includegraphics[width=0.95\textwidth]{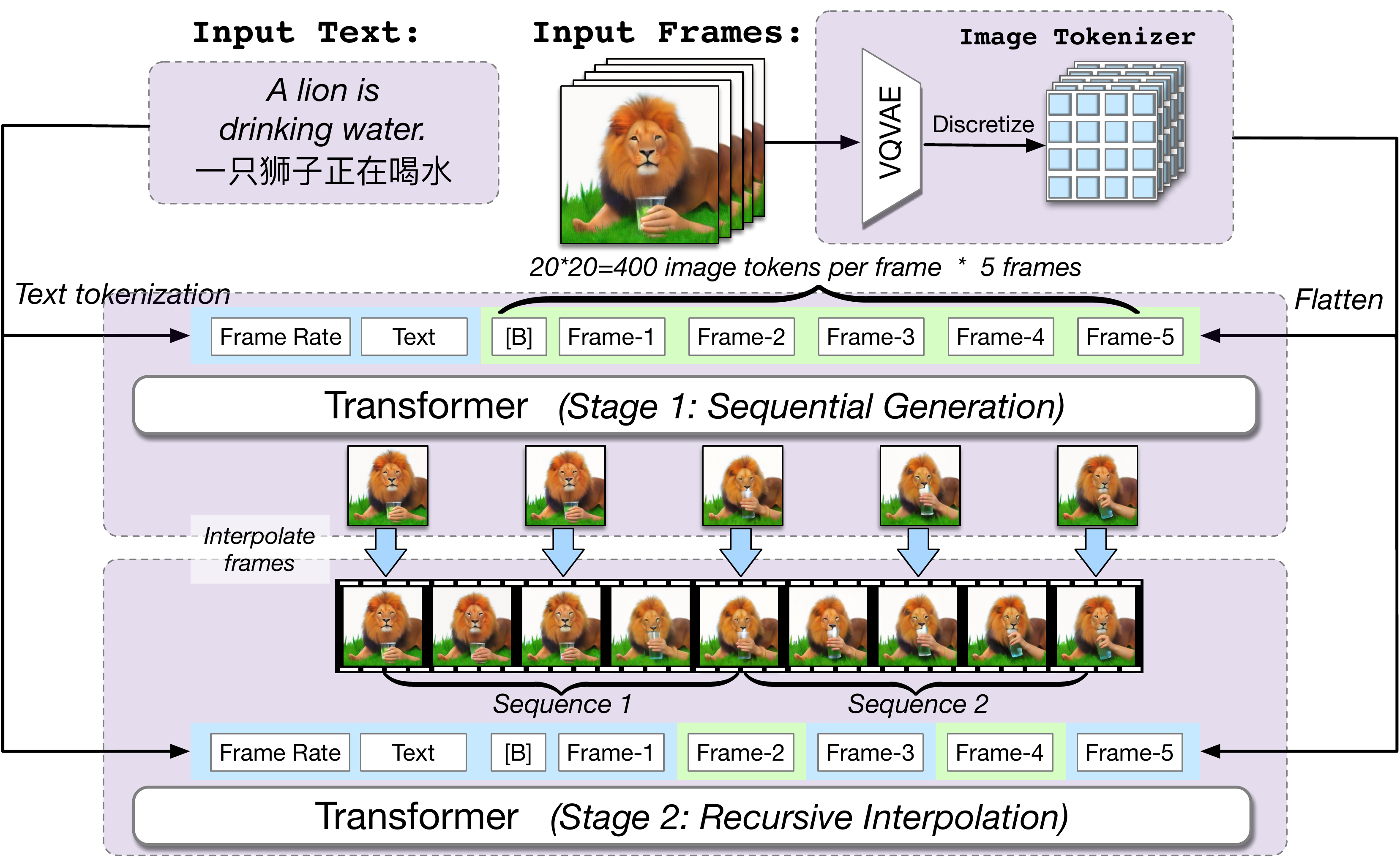}
  \caption{Multi-frame-rate hierarchical generation framework in CogVideo. Input sequence includes frame rate, text, frame tokens. [B] (Begin-of-image) is a separator token, inherited from CogView2. In stage 1, $T_s$ frames are generated sequentially on condition of frame rate and text. Then in stage 2, generated frames are re-input as bidirectional attention regions to recursively interpolate frames. Frame rate can be adjusted during both stages. Bidirectional attention regions are highlighted in \colorbox{MyBlue}{blue}, and unidirectional regions are highlighted in \colorbox{MyGreen}{green}.}
  \label{fig:framework}
\end{figure}

In this section, we first introduce \textit{multi-frame-rate hierarchical training} to better align text and video semantics in \S~\ref{section-2-stage-generation}, and then illustrate an efficient method \textit{dual-channel attention} to inherit the knowledge in pretrained text-image models for video generation in \S~\ref{subsection-2-channel-attention}. To overcome the large memory and time overhead caused by the large model and long sequence, we refer to Swin Attention~\cite{liu2021swin} and extend it to autoregressive video generation in \S~\ref{subsec: swin}.

\subsection{Multi-frame-rate Hierarchical Training} \label{section-2-stage-generation}
Here we present the \textit{multi-frame-rate hierarchical training} and generation. We follow the framework of VQVAE~\cite{van2017neural} and first tokenize each frame into image tokens. Each training sample consists of 5 frames of tokens, but our training method differs in the construction of training sequences and generation process.

\textbf{Training. }The key design is to add a frame-rate token to the text and sample frames at this frame rate to compose a fixed-length training sequence. The motivations are two folds: 
\begin{enumerate}[(1)]
    \item Directly separating the long video into clips at a fixed frame rate often leads to semantic mismatching. We still use the full text but the truncated clip might only contain incomplete action. 
    \item The adjacent frames are usually very similar. A giant change over the previous frame will probably incur a large loss.  This will lead the models less inclined to explore the long-range correlation because simply copying the previous frame acts like a shortcut.
\end{enumerate}

Therefore, in each training sample, we want the text and the frames to match as possible. We predefined a series of frame rates, and select the lowest frame rate for each text-video pair, as long as we can sample at least 5 frames at this frame rate in the video.

Although the above method increases the alignment of text and video, the generation at a low frame rate could be incoherent. We train another \emph{frame interpolation} model to insert transition frames to the generated samples of the sequential generation model. Thanks to the generality of CogLM~\cite{ding2022cogview2}, the two models can share the same structure and training process only with different attention masks. 

\textbf{Generation.} The multi-frame-rate hierarchical generation is a recursive process, illustrated in Figure~\ref{fig:framework}. Specifically, the generation pipeline consists of a sequential generation stage and a recursive interpolation stage:
\begin{enumerate}[(1)]
\item Sequentially generate $T_{s}$ key frames based on a low frame rate and text. The input sequence is \texttt{[\{Frame Rate\}\{Text\} [B] \{Frame$1$\} ...  \{Frame $T_s$\}]}. In practice, we always set $T_{s}=5$ and the minimum sampling frame rate to 1 fps. 

\item Recursively interpolate frames based on the text, frame rate and known frames. In each round of interpolation, we split generated frames into multiple $\lceil \frac{T_s}{2} \rceil$-frame blocks overlapping at the beginning and the end, and interpolate a frame between the successive frames in each block. The input sequence is \texttt{[\{Frame Rate\}\{Text\} [B] \{Frame1\} ... \{Frame $T_s$\}]}, where Frame $2i (i=1,2,...,\lfloor \frac{T_s}{2} \rfloor)$ are to be autoregressively generated. By recursively halfing \texttt{\{Frame Rate\}}, we can conduct finer and finer interpolation to generate videos of many frames.  
\end{enumerate}

\textbf{The effect of CogLM. }Tasks such as frame interpolation rely heavily on bidirectional information. However, most previous works use GPT~\cite{wu2021godiva, yan2021videogpt, wu2021n}, which is unidirectional. To be aware of the bidirectional context, we adopt Cross-Modal General Language Model (CogLM) proposed in~\cite{ding2022cogview2} which unites bidirectional context-aware mask prediction and autoregressive generation by dividing tokens into unidirectional and bidirectional attention regions. While bidirectional regions can attend to all bidirectional regions, unidirectional regions can attend to all bidirectional regions and previous unidirectional regions. As shown in \ref{fig:framework}, (1) all frames in stage 1 and the 2nd, 4th frames in stage 2 are in the unidirectional region; (2) \texttt{\{Frame Rate\}}, \texttt{\{Text\}} and all other frames belong to the bidirectional region. In this way, bidirectional attention context is fully exploited in text and given frames without interfering with auto-regressive frame prediction.

\subsection{Dual-channel Attention} \label{subsection-2-channel-attention}

\begin{wrapfigure}[18]{r}{0.32\textwidth}
  \centering
  \vspace{-5mm}%
  \includegraphics[width=0.32\textwidth]{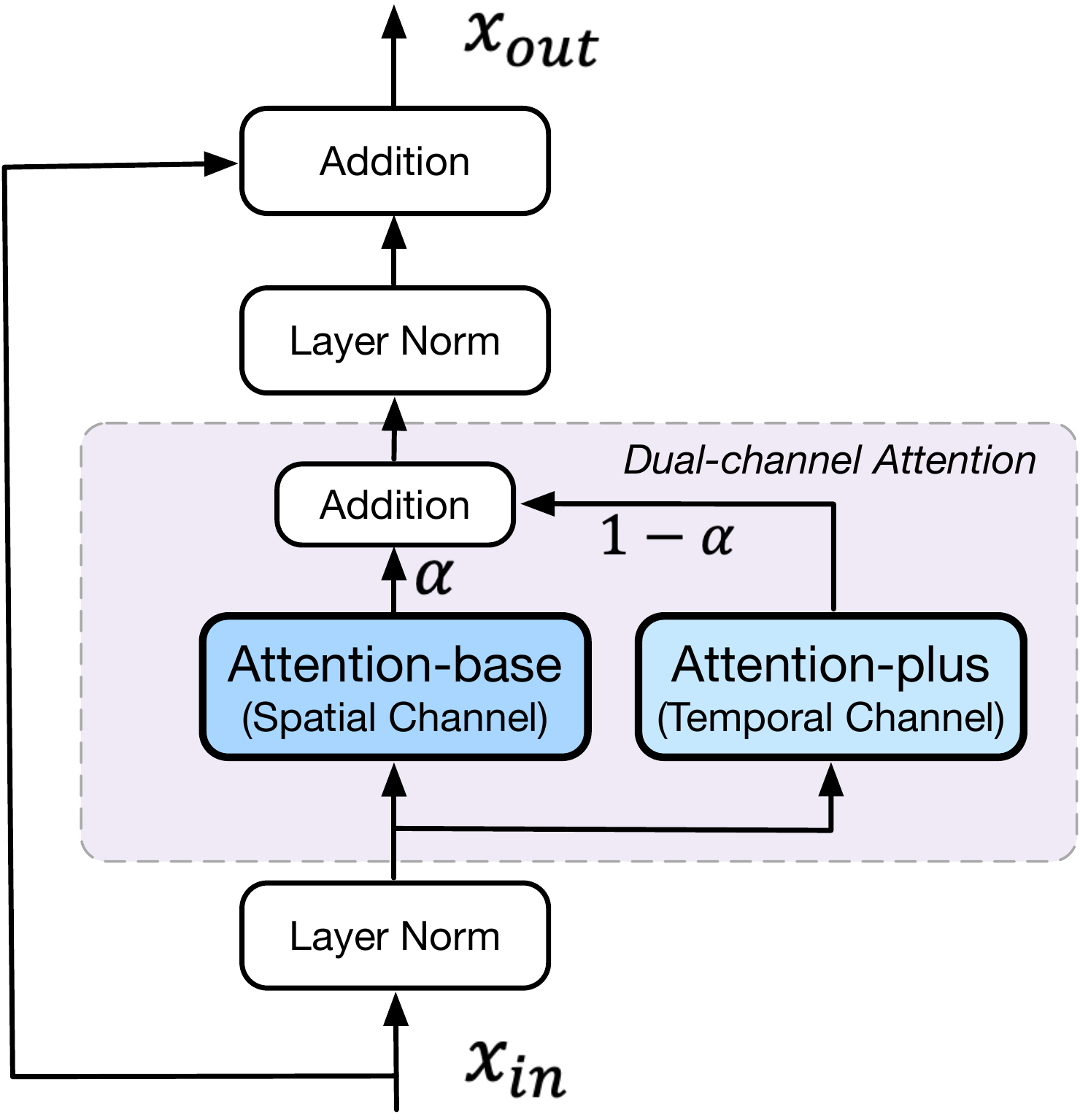}
  \vspace{-6mm}%
  \caption{The dual-channel attention block. We initialize the Attention-plus the same as Attention-base so that the model behaves exactly the same as CogView2 when it is initialized.}
  \label{fig:2-channel-attention}
\end{wrapfigure}

Large-scale pretraining usually demands a large dataset. For the open-domain text-to-video generation, ideally we need the dataset to cover sufficient text-video pairs to infer both spatial and temporal correlation between video and text. However, to collect high-quality text-video pairs is often difficult, expensive and time-consuming.

A natural idea is to make use of the image data to facilitate the learning of spatial semantics.
Video Diffusion Model~\cite{ho2022video} and NÜWA~\cite{wu2021n} try to add text-image pairs into text-video training, which achieves better results on multiple metrics. However, as for training a video-only generation model, adding image data will significantly increase training costs, especially in large-scale pretraining scenarios. 

In this paper, we propose to leverage pretrained image generation models instead of image data. Pretrained text-to-image models, e.g. CogView2~\cite{ding2022cogview2}, already have a good command of the text-image relations. The coverage of the dataset to train these models is also larger than that of videos.  

The proposed technique is \emph{dual-channel attention}, where we only add a new spatial-temporal attention channel to the pretrained CogView2~\cite{ding2022cogview2} at each transformer layer. All the parameters in the CogView2 are frozen in the training, and only the parameters in the newly added attention layer (See the attention-plus in Figure~\ref{fig:2-channel-attention}) are trainable. We denote the original attention block in CogView2 as attention-base. 

Here we also emphasize that directly finetuning CogView2 for text-to-video generation cannot well inherit the knowledge, because the temporal attention follows a different attention pattern and quickly ruins the pretrained weights during the initial phase of training with large gradients.

Specifically, the dual-channel attention block with Sandwich-LN~\cite{ding2021cogview} can be computed as
\begin{align}
\widetilde{x} & = \mathbf{{\alpha}} \cdot \text{attention-base}(\text{LayerNorm}(x_{in}))\notag \\ &\quad+ (\mathbf{1-{\alpha}}) \cdot \text{attention-plus}(\text{LayerNorm}(x_{in})), \label{eq-alpha} \\
x_{out} & = x_{in} + \text{LayerNorm}(\widetilde{x}).
\end{align}
The mixture factor $\alpha$ is a vector $\in {(0,1)}^{d}$, where $d$ is the hidden size of the input feature $x_{in}$. To restrict the range of $\alpha$ within $(0,1)$, we reparameterize it as $\alpha = \text{sigmoid}(\mathbf{a}) \in {(0,1)}^{d}$, where $\mathbf{a}\in \mathbb{R}^d$ is a learnable parameter. 
The attention-plus block has the same shape of parameters as the normal multi-head attention block, attention-base, but differs in the  procedure of computation as follows.

In our training, we tried two kinds of attention, 3D local attention and 3D Swin~\cite{liu2021swin} attention for attention-plus block. Here we depict the 3D local attention, and the latter is a natural replacement introduced in section~\ref{subsec: swin}.

In 3D local attention, the receptive field (RF) for the token at $(t, x, y)$ 
(where $(t, x, y)$ corresponds to the coordination along time, height and width), is a 3D block with extent $l_t, l_x, l_y \in \mathbb{N}^+$:
\begin{equation}
    \text{RF}_{(t, x, y)} = \{(k,i,j) ~ \Big\vert ~ \lvert x-i \rvert < l_x, \lvert y-j \rvert < l_y,  \lvert t-k \rvert < l_t, ~(k, i, j) \notin \text{Mask}_{(t, x, y)} \}, 
\end{equation}
where $\text{Mask}_{(t, x, y)}$ represents an attention mask for token $(t, x, y)$. In the sequential generation model (Stage 1), the Mask ensures the auto-regressive order; In the interpolation model (Stage 2), the Mask is designed as in as CogLM~\cite{ding2022cogview2} to make the known frames visible to all the frames. 

It is worth noting that two channels are fused and share the same FFN in each layer, 
because FFN is a module of heavy parameters containing much vision knowledge. Due to the similarity between images and videos, bringing its knowledge to the temporal channel will facilitate video modeling. Finally, sharing FFN can reduce parameters, thus speeding up training and reducing memory overhead.

\subsection{Shifted Window Attention in Auto-regressive Generation} \label{subsec: swin}
To further alleviate the large time and memory overhead in the temporal channel during training and inference, we refer to Swin Attention~\cite{liu2021swin}. 
The original Swin attention is only applied to non-autoregressive scenarios, we extend it to the autoregressive and temporal scenario by applying an auto-regressive attention mask in the shifted windows. 

\begin{wrapfigure}[15]{r}{0.5\textwidth}
  \centering
  \vspace{-5mm}%
  \includegraphics[width=0.5\textwidth]{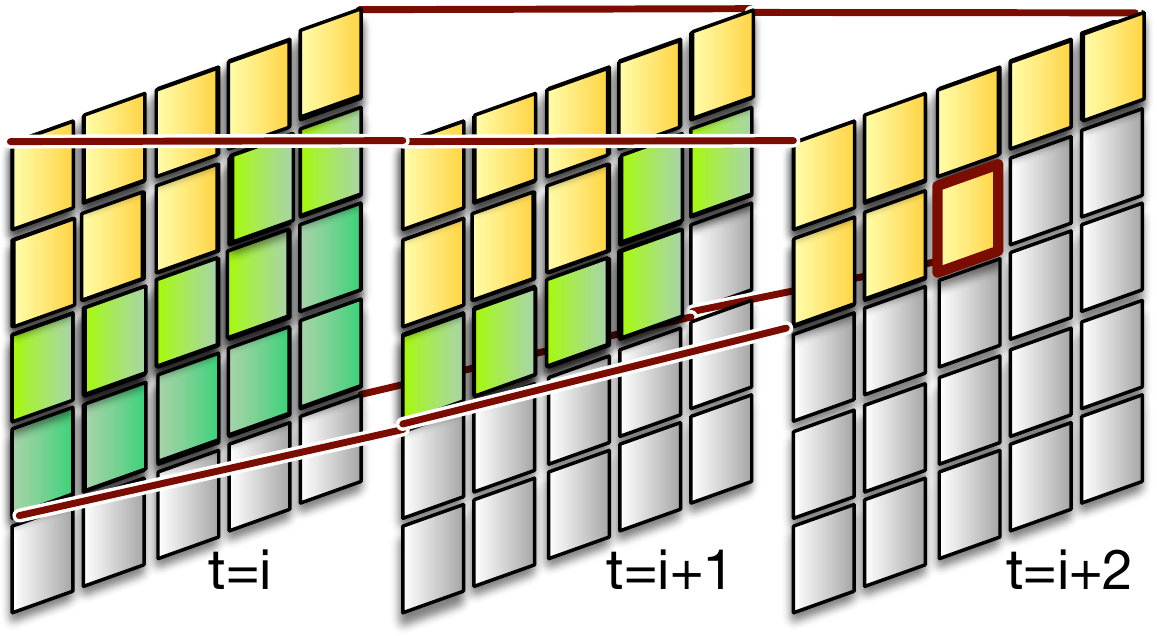}
  \vspace{-4mm}%
  \caption{In 3D autoregressive swin attention (window size $2\times2$ as an example), the token in the red box can only attend to (either directly or indirectly) the yellow or green tokens. The gray tokens in the $i$-th frame and the token in the red box can be generated in parallel.}
  \label{fig:swin}
\end{wrapfigure}

An interesting finding is that, \textbf{the Swin attention provide a chance for parallel generation in faraway regions of different frames}, which further accelerates the auto-regressive generation.  
The dependence of the generation of a specific token relies on
\begin{itemize}
    \item Auto-regressive mask. A token can only attend to previous frames or tokens before itself in the current frame. 
    \item Shifted window. Only tokens within the distance of window size in both width and height dimensions can be directly attended to.
\end{itemize}
As shown in Figure~\ref{fig:swin}, we can start generating parts of the tokens in the following frames before finishing the generation of all the previous frames --- they can work in parallel. 
Suppose $X$,$Y$ is the height and width of each frame, and $A_x$,$A_y$ are the height and width of shifted window. For two tokens at $(t_1, x_1, y_1)$ and $(t_2, x_2, y_2)$, $t_1 < t_2$, the latter cannot attend to the former either directly or indirectly if
\begin{equation}
    (x_1-x_2)Y+(y_1-y_2) \geq (t_2-t_1+1)(A_xY+A_y),
\end{equation}which means that the $i$-th token in the $t$-th frame can be generated with the $(i-A_xY-A_y)$-th token in the $(t+1)$-th frame in parallel. In this way, we can generate $\lfloor \frac{XY}{A_xY+A_y} \rfloor$ tokens in parallel at most, thus greatly enhance parallelism and accelerate inference compared to auto-regressive with standard attention which can only generate one token at a time. 

\section{Training}

Based on the methods above, the training details of CogVideo are listed as follows: 

\textbf{Model. }The backbone of CogVideo in both stages is a Transformer with dual-channel attention. The Transformer has 48 layers, with a hidden size of 3,072 in each attention channel, 48 attention heads and 9.4 billion parameters in total. Among them, 6 billion parameters are fixed to CogView2's parameters, which include Position-wise Feed-Forward Networks (FFN), the spatial channel of dual-channel attention, first frame's positional embeddings and all image and text vocabulary embeddings. The specific implementation of Transformer structure is almost identical to CogView~\cite{ding2021cogview} such as using Sandwich LayerNorm and PB-Relax to stabilize training. Shifted CogLM attention window is adopted in recursive interpolation model with window size $10 \times 10$. 

\textbf{Dataset. }We pretrain our model on a dataset of 5.4 million captioned videos with a spatial resolution of $160\times 160$ (can be upsampled to $480\times 480$ by CogView2). For the sequential generation model (Stage 1), we adjust the frame rate in each sample to accommodate the whole video, while the minimum frame rate is set to 1 fps. For the recursive interpolation model (Stage 2), we split videos into clips of different lengths to accommodate prediction on multiple frame rates including 2,4,8 fps. 

\textbf{Pretraining. }The sequence lengths in both stages are 2,065, consisting of 64 text tokens, 5 (frames) $\times$ 400 (per frame)  image tokens, and 1 seperator token. Both text and images are tokenized with icetk\footnote{\url{https://github.com/THUDM/icetk}}.The parameters are updated by Adam with max learning rate $= 2\times10^{-4}$, $\beta_1 = 0.9$, $\beta_2 = 0.95$, weight decay $=1\times10^{-2}$. See Appendix for pretraining details. 

\section{Experiments} \label{sec-experiment}


\subsection{Machine Evaluation}
Machine evaluation is conducted on two popular benchmarks for video generation, i.e., UCF101~\cite{soomro2012ucf101} and Kinetics-600~\cite{carreira2018short}. Following~\citet{rakhimov2020latent,yu2022generating}, we use 
Fréchet Video Distance (FVD)~\cite{unterthiner2018towards} and Inception score (IS)~\cite{salimans2016improved} as metrics in the evaluation. FVD is calculated based on I3D model\cite{carreira2017quo} trained on Kinetics-400, and IS is based on C3D model~\cite{tran2015learning} which was first trained on the Sports-1M dataset~\cite{karpathy2014large} and then finetuned on the UCF101 dataset. Our evaluation code is the same as the official TGAN-v2 implementation\footnote{\url{https://github.com/pfnet-research/tgan2}}. 

\textbf{UCF-101} is a human action dataset consisting of 13,320 videos 
annotated 
with 101 action classes. Due to the gaps of image style and frame rate between CogVideo's training set and UCF-101, we use class labels as the input text and finetune CogVideo on the whole dataset for 10,000 iterations with a batch size of 192. During inference, we generate samples of various classes according to the class distribution.
FVD and IS are evaluated over 2,048 and 10,000 samples respectively, following \citet{yu2022generating}. Results are shown in Table~\ref{tab:machine eval} (Left). 


\begin{table}
    \caption{(Left) Video generation performance on UCF-101. Class labels are used as the text inputs. * means that the model is only trained on the training split of UCF-101. (Right) Video generation performance on Kinetics-600. The metrics are based on the 16-frame generated videos priming on first 5 frames, following settings of  \citet{rakhimov2020latent}. ** means that the ground truth used in FVD testing is the reconstruction result of the tokenizer.}
  \label{tab:machine eval}
  \centering
  \begin{minipage}{0.47\columnwidth}
  \centering
  \begin{tabular}{lcc}
    \toprule
    Method & IS ($\uparrow$) & FVD ($\downarrow$)\\
    \midrule
    VideoGPT\cite{yan2021videogpt}  &  24.69 & - \\
    DVD-GAN\cite{clark2019adversarial} & 27.38 & - \\
    TGANv2\cite{saito2020train}* & 28.87 & 1209 \\
    MoCoGAN-HD\cite{tian2021good} & 32.36 & 838  \\
    DIGAN\cite{yu2022generating}* & 29.71  & 655 \\
    DIGAN\cite{yu2022generating} & 32.70  & 577\\
    TATS-base\cite{ge2022long} & 79.28  & 332\\
    \midrule
    CogVideo (Ours) & 50.46  & 626\\
    CogVideo (Ours)** & - & 545 \\
    \bottomrule
  \end{tabular}
\end{minipage}
\begin{minipage}{0.47\columnwidth}
  \centering
  \begin{tabular}{lc}
    \toprule
    Method & FVD($\downarrow$) \\
    \midrule
    Latent Video Tranformer\cite{rakhimov2020latent} & 224.73    \\
    Video Transformer\cite{weissenborn2019scaling}     & 170 \\
    DVD-GAN-FP\cite{clark2019adversarial}     & 69.15  \\
    TriVD-GAN-FP\cite{luc2020transformation}  & 25.74  \\
    \midrule
    CogVideo (Ours) & 109.23 \\
    CogVideo (Ours)** & 59.55 \\
    \bottomrule
  \end{tabular}
  \end{minipage}
\end{table}

\textbf{Kinetics-600} contains 600 classes of human action videos, with roughly 350,000 train and 50,000 test videos in total. We use the action category as input text, and finetune CogVideo on the training set for 12,000 iterations with a batch size of 640. Following the setup of \citet{weissenborn2019scaling, rakhimov2020latent}, we center-crop and downsample each frame to 64$\times$64 to measure the 
FVD of the model. Results are shown in Table~\ref{tab:machine eval} (Right). 

\subsection{Human Evaluation} \label{subsec:human}
\begin{figure}
  \centering
  \includegraphics[width=\textwidth]{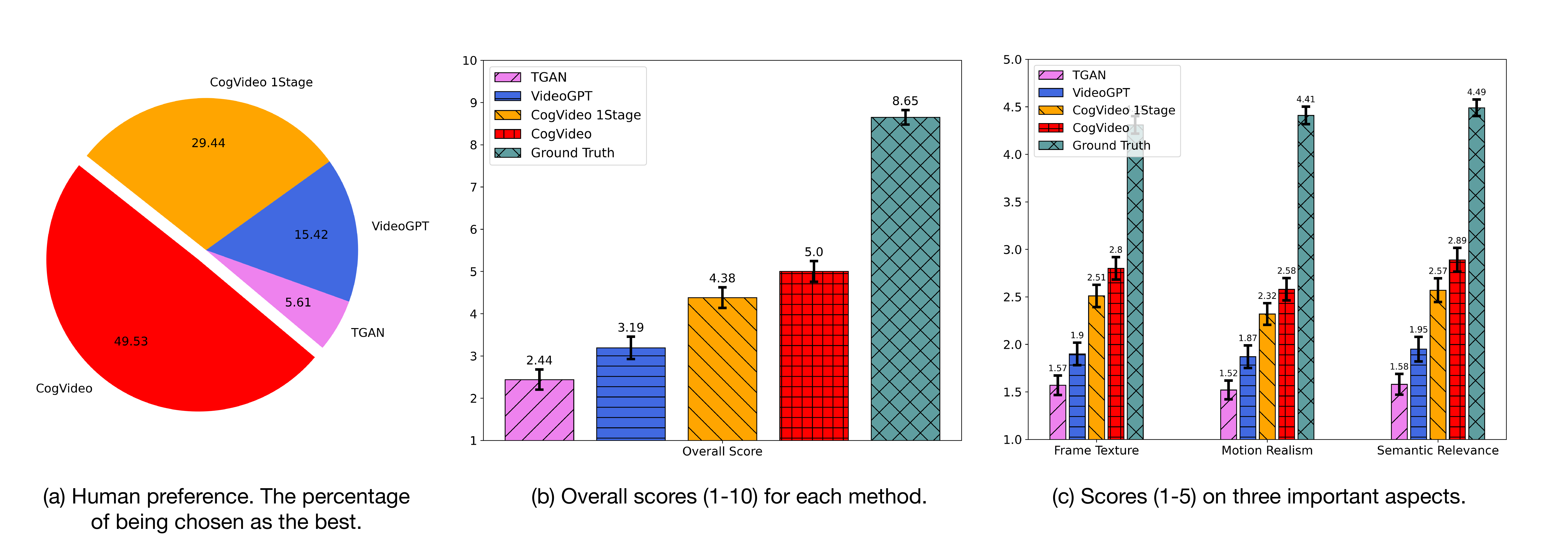}
  \caption{Human evaluation results. ``CogVideo 1Stage'' refers to the method in ablation study, which only generates videos sequentially with the CogVideo's Stage 1 to the desired number of frames. }
  \label{fig:human-evaluation}
  \vspace{-4mm}
\end{figure}

To further evaluate CogVideo, we invite 90 anonymous evaluators to rate for CogVideo and other open-source baselines including GAN-based model TGANv2~\cite{saito2020train} and GPT-based model VideoGPT~\cite{yan2021videogpt}. 30 classes in UCF101 are randomly picked as text conditions, and several aspects are rated (See Appendix for details). For VideoGPT, we use the official unconditional pretrained model\footnote{\url{https://github.com/wilson1yan/VideoGPT}} to generate samples. For TGANv2, we use the official source code to train an unconditional generation model under the same setting as that in~\citet{saito2020train}. To assign unconditionally generated samples into corresponding categories, we choose TSM~\cite{lin2019tsm} as the action recognition model for a post-classification. We only keep the samples whose likelihood to a certain class is at least 80\%.
Results in Figure~\ref{fig:human-evaluation} show that CogVideo 
significantly outperforms baselines
on multiple important aspects including frame texture, motion realism and semantic relevance, and achieves the top score by the overall quality. 
It can be seen that
49.53\% evaluators choose CogVideo as the best method, and only 15.42\% and 5.6\% favor VideoGPT and TGANv2, respectively.

\subsection{Ablation Study} \label{subsec:ablation study}
To verify the effectiveness of hierarchical multi-frame-rate generation and incorporating CogView2, we conduct ablation studies on Kinetics-600 and UCF-101 datasets. We will first briefly introduce the compared methods and analyze the quantitative results in \S~\ref{sec:ab1} and qualitative results in \S~\ref{sec:ab2}

\textbf{Hierarchical multi-frame-rate generation.} In comparison with CogVideo, we finetune a 1-stage video generation model on Kinetics-600 from the sequential generation model in CogVideo, which generates long videos by sliding windows. In each window, we generate the rest frames based on $N_{overlap}$ previous known frames. Larger $N_{overlap}$ means more previous frames can be utilized during the inference, but will increase time overhead. 

\textbf{Dual-channel attention with CogView2's weights.} To highlight the effectiveness of our finetuning strategy, we additionally finetune (1) a randomly initialized model, (2) a model incorporating CogView2's weights but leaving the temporal channel unfixed (equivalent to CogVideo without pretraining on videos) on Kinetics-600 for comparison.

\subsubsection{Quantitative Evaluation}\label{sec:ab1}
\begin{table}
  \caption{Ablation study on a 5,000-sample subset of Kinetics-600's testset. FVD is evaluated on generated 11-frame samples priming on 5 frames and the recovered ground-truth by the image tokenizer. The setting column indicates the difference between each method and CogVideo. Models of each setting are trained on Kinetics-600 trainset for 11,000 iterations with a batch size of 160. }
  \label{tab:ablation-study}
  \centering
  \begin{tabular}{lcc}
    \toprule
    Method & Setting & FVD ($\downarrow$)              \\
    \midrule
    CogVideo & None & 108.27\\
    \midrule
    1-stage Generation($N_{overlap}=1$) & $-$ hierarchical & 137.13\\
    1-stage Generation($N_{overlap}=2$) & $-$ hierarchical & 120.82\\
    \midrule
    Initialized with CogView2 & $-$ Pretrain & 124.92 \\
    Randomly Initialized & $-$ Pretrain $-$ CogView2 & 166.13 \\
    
    \bottomrule
  \end{tabular}
\end{table}

\begin{wrapfigure}[10]{r}{0.4\textwidth}
  \centering
  \vspace{-12mm}%
  \includegraphics[width=0.4\textwidth]{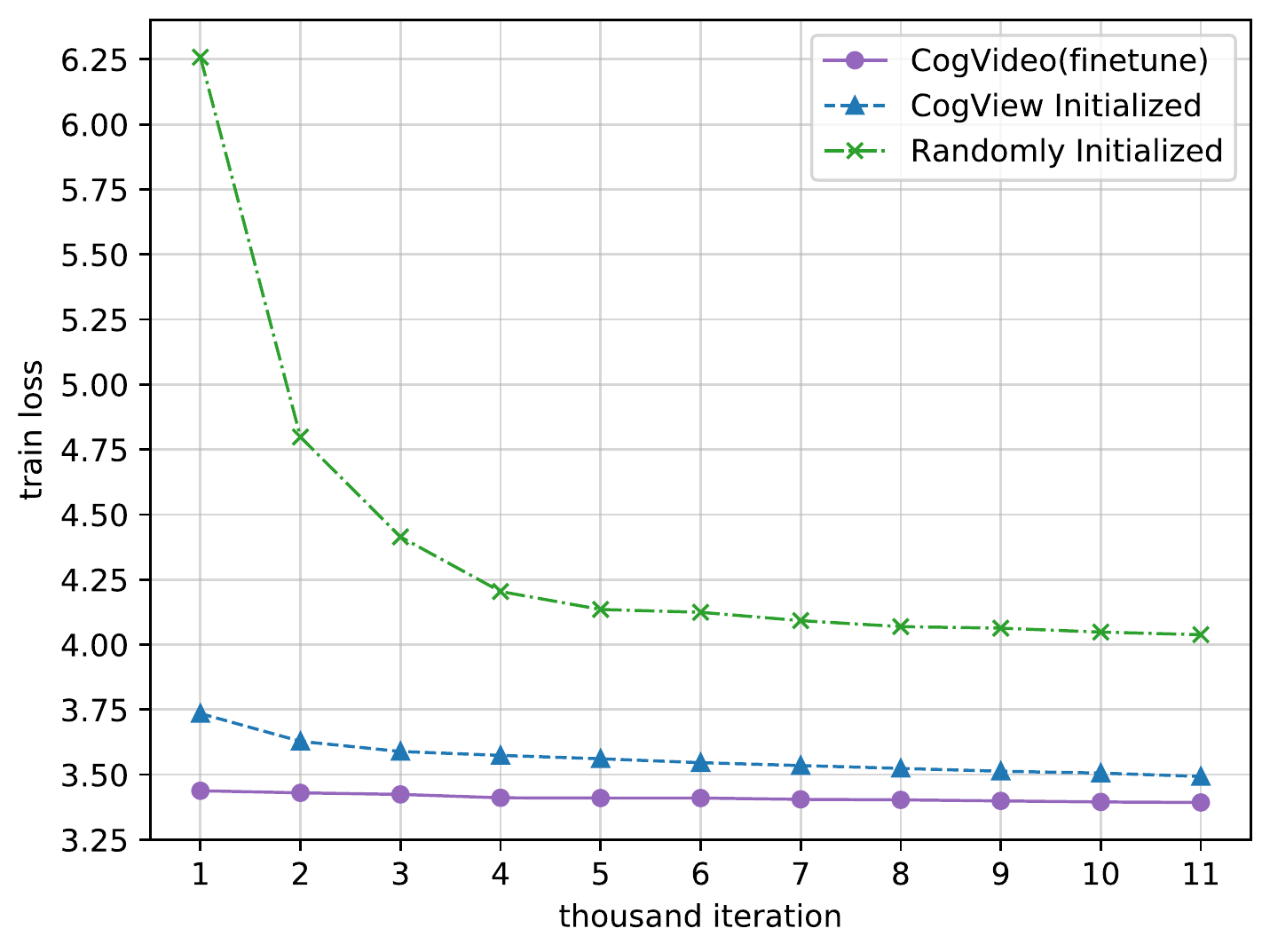}
    \vspace{-7mm}%
  \caption{Training loss in ablation study. }
  \label{fig:ablation-loss}
\end{wrapfigure}

All aforementioned models 
have been trained for 11,000 iterations with a batch size of 160. Quantitative 
results are shown in Table~\ref{tab:ablation-study}. 
We can see that the hierarchical method is clearly superior to the 1-stage generation with different $N_{overlap}$, and the model initialized with CogView2's weights has lower FVD than the randomly initialized one. 

Figure~\ref{fig:ablation-loss} plots the training loss curve of (1) finetuning CogVideo; (2) training model from random initialization; (3) training model initialized with CogView2 and partially fixed. We can see that CogView2 endows the model with a good initialization point from which the loss can decrease faster. Moreover, fixing part of the parameters reduces the time and memory cost.

\subsubsection{Qualitative Evaluation}\label{sec:ab2}

\begin{figure}
\centering
  \includegraphics[width=\textwidth]{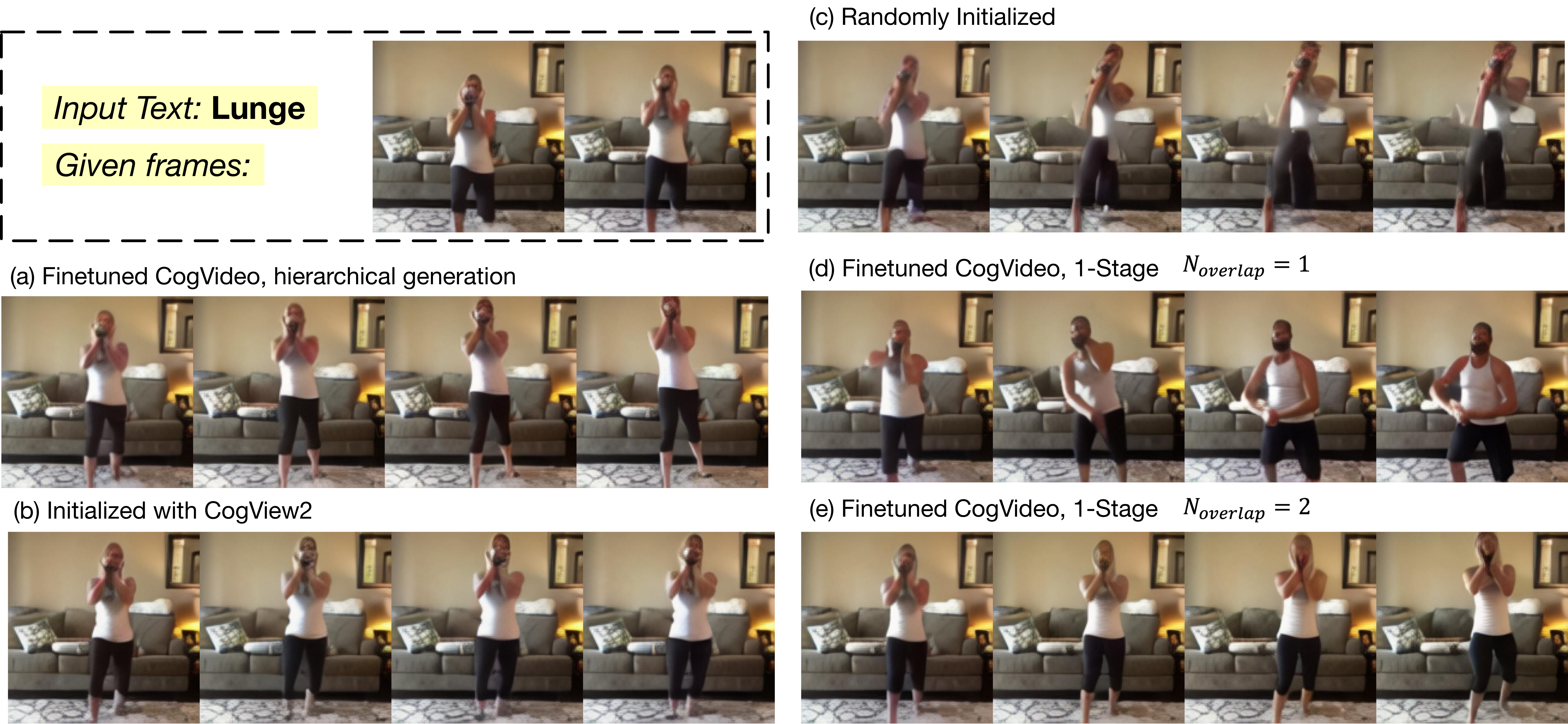}
  \vspace{-4mm}%
  \caption{Video samples in ablation study, which are generated priming on the class label and first 5 frames in Kinetics-600. All samples are downsampled by extracting one in every three frames for display purposes. (a) Use finetuned CogVideo to hierarchically generate samples. (b) Train a model on Kinetics-600 which is initialized as and partially fixed to CogView2, and hierarchically generate samples. (c) Train a model on Kinetics-600 which is randomly initialized, and hierarchically generate samples. (d)(e) Use finetuned CogVideo to generate frames in 1 stage with different $N_{overlap}$. }
  \label{fig:ablation-case}
\end{figure}

Qualitative comparison is shown in Figure~\ref{fig:ablation-case}. While the model trained from random initialization tends to produce irrational deformation, the model incorporating CogView2 is able to generate realistic objects, and the hierarchical generation performs better on content consistency and motion realism. 

We also conduct human evaluation between 1-stage and hierarchical video generation model under the same setting as in \S~\ref{subsec:human}. As shown in Figure \ref{fig:human-evaluation}, the hierarchical model, i.e. CogVideo, outperforms the 1-stage model on semantic relevance, motion realism as well as texture quality. 
This is probably because the 1-stage model cannot estimate a proper intensity of change from the previous frames in the window, as shown in Figure~\ref{fig:ablation-case}(d)(e). 

\section{Conclusion}
We present CogVideo, to the best of our knowledge, 
the largest and the first open-source pretrained transformer for text-to-video generation in general domain. CogVideo is also the first attempt to efficiently leverage the pretrained text-to-image generative model to the text-to-video generation model without hurting its image generation capacity.
With the proposed multi-frame-rate hierarchical training framework, CogVideo is endowed with a better understanding of text-video relations and abilities to control the intensity of changes during generation. We extend swin attention to CogLM, which achieves acceleration in both training and inference. There are still some limitations in CogVideo, e.g. restriction on the length of the input sequence still exists due to the large scale of the model and limitation of GPU memory, and we leave them for future work. 

\textbf{Broader Impact. }This paper aims to advance the open-domain text-to-video generation, which will ease the effort of short video and digital art creation. The efficient training method transfers knowledge from text-to-image models to text-to-video models, which helps avoid training from scratch, and thus reduces energy consumption and carbon emission. A negative impact is the risk of misinformation. To alleviate it, we can train an additional classifier to discriminate the fakes. We believe the benefits outweigh the downsides.

\begin{ack}
We would like to thank Zhao Xue, Shuai Zhao, Sha Yuan for their help in data collection, Weidong Guo, Fengyu Rao, Zhaoyang Zeng, Mingkang Tang for their useful discussion, Hanxiao Qu for maintaining the machines and the computational resources supported by BAAI.  
\end{ack}



\bibliography{neurips_2022} 
\bibliographystyle{abbrvnat}

\appendix

\section{Attention Analysis}
To explore the attention mechanism of \emph{dual-channel attention}, we visualize (1) the attention distribution in the temporal channel and (2) the mixture factor $\alpha$ controlling the ratio between the spatial and temporal channel in equation~\ref{eq-alpha}. 

\begin{figure}[ht]
  \centering
  \includegraphics[width=\textwidth]{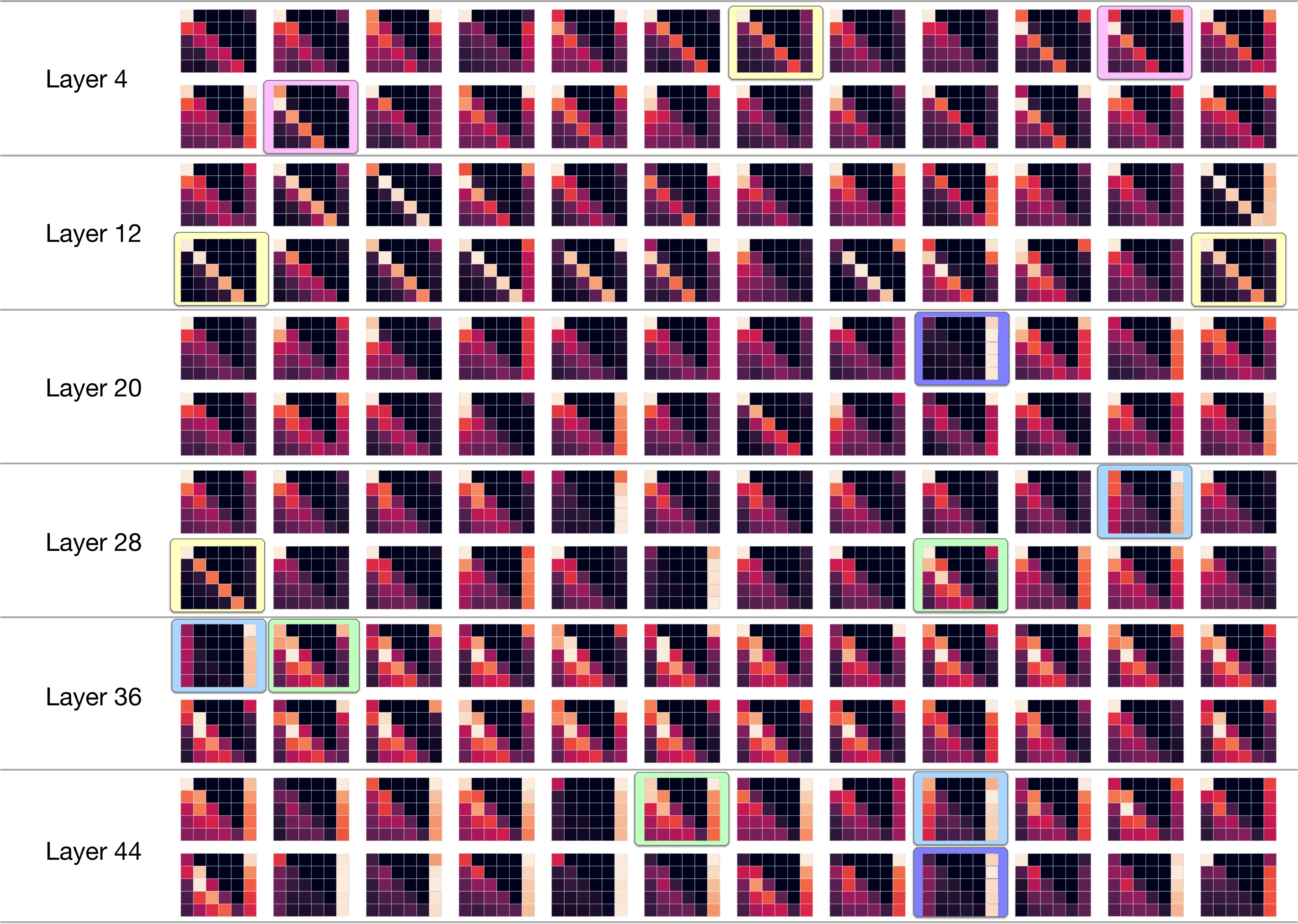}
  \vspace{-5mm}
  \caption{The attention distribution among frames and texts in sequential generation (Stage 1). Only 24 of 48 attention heads in 6 layers are selected for display purposes. Each attention head is visualized with a heat map of size 5$\times$6, where lighter color represents larger value. The 5$\times$5 block on the left indicates the sum of attention scores (after softmax) between each pair of frames, and the rightmost column indicates the sum of the attention score of each frame to text. That is to say, the grid in row i column j ($j\leq 5$) represents $\sum_{x\in F_i, y\in F_j} \text{attn}_{x, y}$, and the grid in row i column 6 represents $\sum_{x\in F_i, y\in T} \text{attn}_{x, y}$, where $F_i$, $T$ denotes the set of tokens in the i-th frame and text respectively, and $\text{attn}_{x, y}$ denotes the attention score of token x to y. }
  \label{fig:appendix-heatmap}
\end{figure}

Figure~\ref{fig:appendix-heatmap} visualizes the distribution among frames and texts in sequential generation (Stage 1) with heat maps, where only 24 of 48 attention heads in 6 layers are shown for display purposes. The attention patterns can be broadly classified into the following categories: 
\begin{itemize}
    \item Most of the attention is on the text. E.g. the attention heads in \colorbox{AppendixViolet}{violet}. 
    \item Most of the attention is on a certain frame. E.g. the attention heads in \colorbox{AppendixPink}{pink} focus mainly on the previous frame; the attention heads in \colorbox{AppendixBlue}{blue} focus mainly on the first frame besides the text; the attention heads in \colorbox{AppendixYellow}{yellow} focus mostly on the frame itself. 
    \item Attention is spread over several frames. E.g. the attention heads in \colorbox{AppendixGreen}{green}.  
\end{itemize}

Some attention heads exhibit a single pattern, while others may exhibit a mixture of them. Attention heads in the same layer tend to show similar patterns. In lower layers (e.g. layer 4, 12) the heads tend to allocate attention according to position, while in higher layers more attention is allocated to text (e.g. layer 44) or spread over multiple frames. One possible explanation is that there are more high-level features in higher layers such as video semantics, by which more frames and texts can interact with each other to make high-level feature analysis.

\begin{figure}
  \centering
  \includegraphics[width=\textwidth]{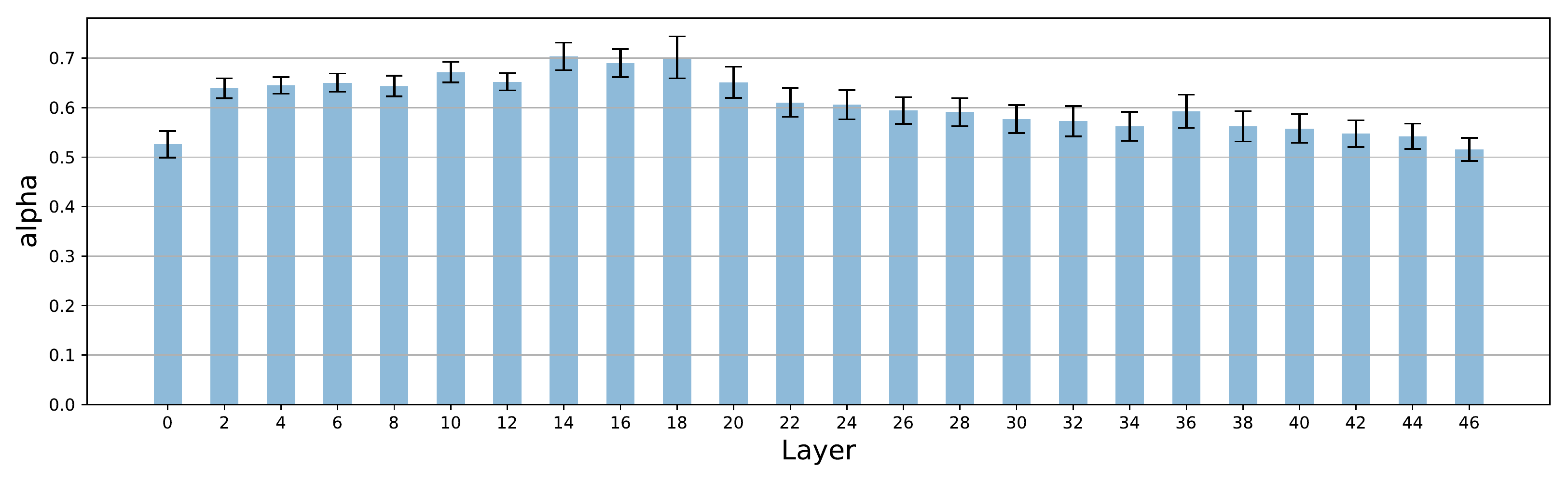}
  \vspace{-7mm}
  \caption{The scale factor $\alpha$ controlling the ratio between the spatial and temporal channel in equation~\ref{eq-alpha} in dual-channel attention. Only $\alpha$ in half of the layers are shown for display reasons. As $\alpha$ is a vector of dimension 3072, we show the mean and variance among all of its dimensions in this figure. }
  \label{fig:appendix-attrib}
\end{figure}

It is worth noting that many heads in temporal channel do not allocate much attention to the frame itself, especially in higher layers, while attending to itself is important for inference. This shows that the CogVideo performs a certain degree of decoupling in the analysis of temporal and spatial features. While the spatial channel is in charge of feature analysis within the frame, the temporal channel can allocate more resources to explore relationships among different frames. We further illustrate this perspective with Figure~\ref{fig:appendix-attrib}, which shows that features calculated by CogView2 in the spatial channel are heavily relied on. 





\section{Training Details}
CogVideo consists of two models corresponding to two stages, i.e. sequential generation and recursive interpolation. Both models have 7.7 billion parameters while 6 billion of them are fixed to CogView2, thus CogVideo has 9.4 billion different parameters in total. 

CogVideo is trained on a dataset of 5.4 million captioned videos with a spatial resolution of 160$\times$160 (can be upsampled to 480$\times$480 by CogView2). Each model is pretrained separately. The model in stage 1 is first pretrained for 76,000 iterations on video clips with a minimum frame rate of 0.25 fps, then trained for 15,000 iterations with a minimum frame rate of 1 fps. The model in stage 2 is pretrained for 78,500 iterations with the frame rate of 2, 4, and 8 fps. 
Both models are trained in FP16 with a batch size of 416, and optimized by Adam with max learning rate $= 2\times10^{-4}$, $\beta_1 = 0.9$, $\beta_2 = 0.95$, weight decay $=1\times10^{-2}$.

\section{Details about Human Evaluation}
In this section, we introduce more details about the human evaluation for measuring generation quality. The conduction of our human evaluation generally follows previous works including \citet{ramesh2021zero, ding2021cogview}

\begin{figure}
  \centering
  \includegraphics[width=\textwidth]{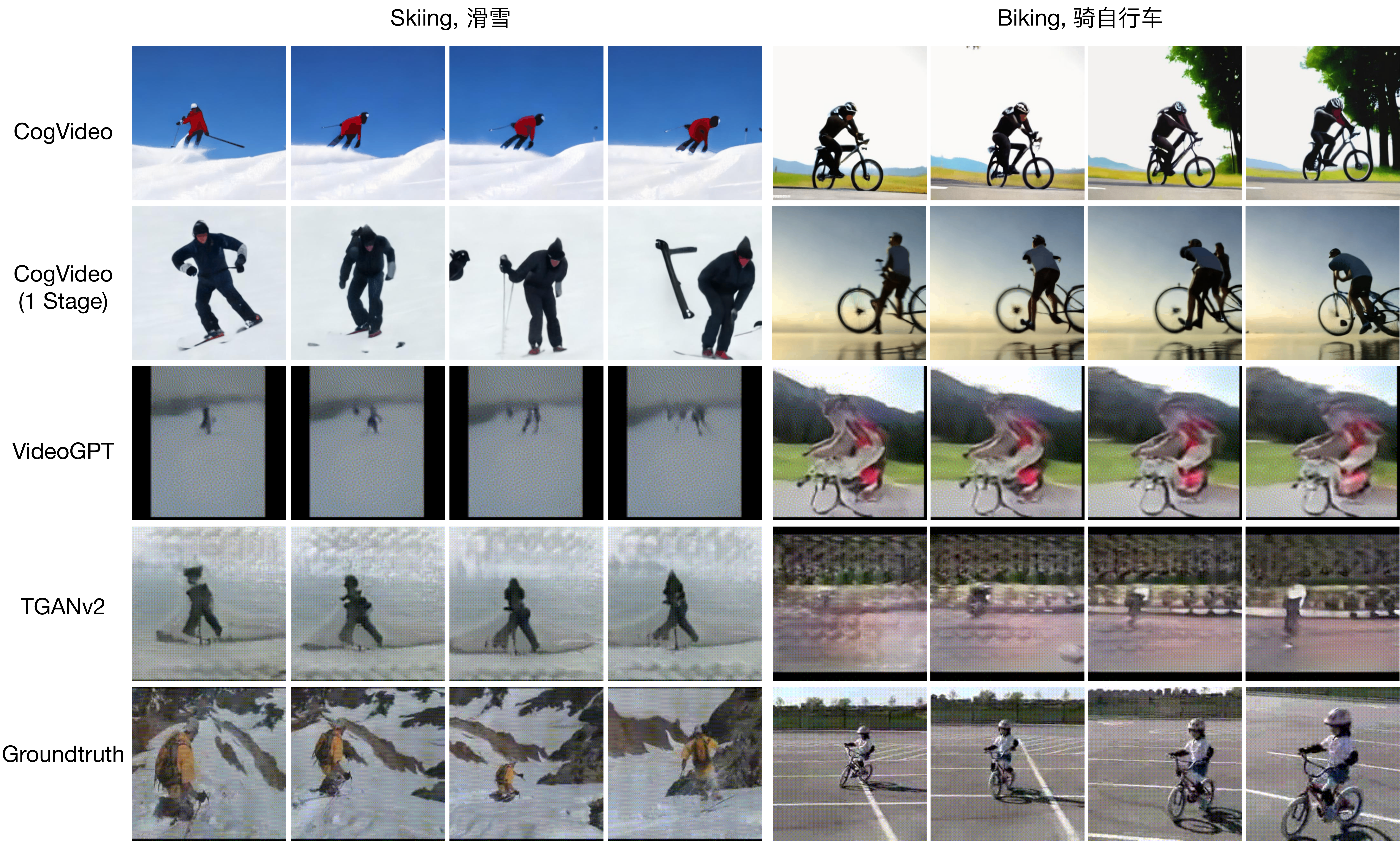}
  \vspace{-4mm}
  \caption{A subset of human evaluation samples. The captions are randomly selected from UCF-101. The original samples are clips of 16 frames, which are downsampled to 4 frames uniformly for display purposes. }
  \label{fig:appendix-humansamples}
\end{figure}

We randomly extract 30 classes from UCF101 for video generation, using corresponding video samples in the dataset as ground truth items in the evaluation. Based on captions of selected classes, we generate video samples from models including TGANv2, VideoGPT, and our model, CogVideo. To further illustrate the effectiveness of hierarchical multi-frame-rate generation, we also include a 1-stage version of CogVideo model fine-tuned on Kinetics-600 which is described in \S~\ref{subsec:ablation study}. For TGANv2, we use the official source code to train an unconditional generation model under the same setting as that in \citet{saito2020train}. For VideoGPT, we use the official unconditional pretrained model to generate samples. To assign unconditionally generated samples into corresponding categories, we choose TSM\cite{lin2019tsm} as the action recognition model for a post-classification. We only keep the samples whose likelihood to a certain class is at least 80\%. A randomly selected subset of samples is displayed in Figure~\ref{fig:appendix-humansamples}. 

\begin{figure}
  \centering
  \includegraphics[width=\textwidth]{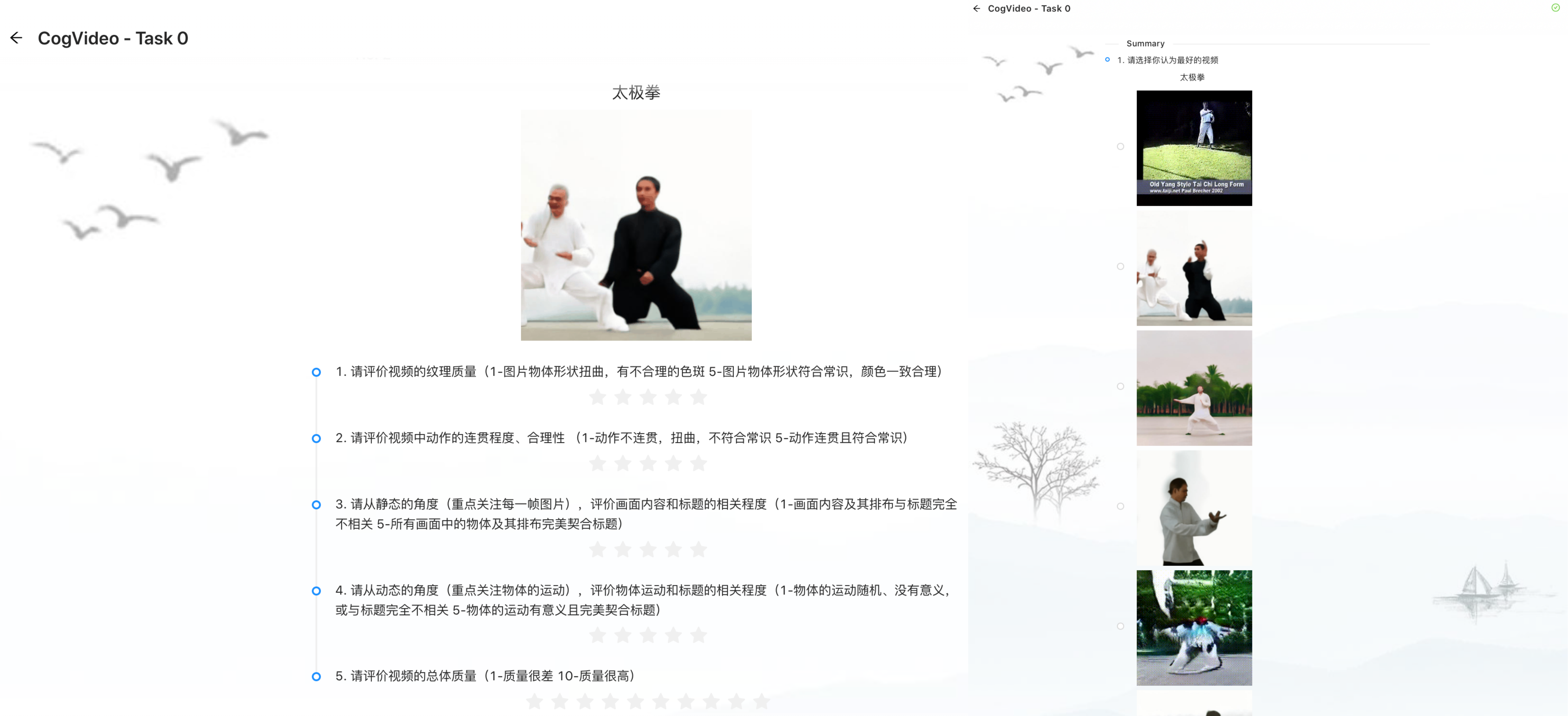}
  \vspace{-4mm}
  \caption{Snapshots of the evaluation website. }
  \label{fig:appendix-snapshot}
\end{figure}

For each sample of the video mentioned above, we ask evaluators to give scores between 1 and 5 ( 5 indicates the best while 1 indicates the worst) from three aspects including frame texture, motion realism, and semantic relevance. Then the evaluators are required to give a general score of quality for each sample between 1 and 10, where a higher score indicates better quality. After video samples from each caption are all evaluated, the evaluators are asked to select the best one from them. We show snapshots of the evaluation website in Figure~\ref{fig:appendix-snapshot}

Throughout the process of human evaluation, we invited nearly 100 anonymous evaluators, while 90 of them completed the whole evaluation and were counted in the final results. None of the questions in the evaluation have any time limit. We offer each evaluator 75 RMB as a reward for the evaluation. Results of the human evaluation, including the average score and standard deviation for each group, have already been introduced in Figure~\ref{fig:human-evaluation} in the main body. As ground truth samples take an absolute predominance in the best selection question, we have removed the part of ground truth samples in the selection pie plot for clearer model comparison.



\end{document}